\documentclass[12pt,a4paper]{article}
\usepackage{newtxtext}
\usepackage{listings}
\usepackage[utf8]{inputenc}
\usepackage{setspace}
\usepackage{amsmath}
\usepackage{amssymb}
\usepackage{latexsym}
\usepackage{graphicx}
\usepackage{url}
\usepackage{ascmac}
\usepackage{amsthm}
\usepackage{multicol}
\usepackage{braket}
\usepackage{enumerate}
\usepackage[super]{nth}
\usepackage{mathtools}
\usepackage{enumerate}
\usepackage{comment}
\usepackage{cancel}
\usepackage{color}
\usepackage{nccmath}
\usepackage{bm}
\usepackage{algorithm}
\usepackage{algpseudocode}
\usepackage{algorithmicx}
\usepackage{xcolor,colortbl,array}
\usepackage{subcaption}
\usepackage{booktabs}
\usepackage{fancyhdr}
\fancypagestyle{plain}{
  \fancyhf{} 
  \fancyfoot[C]{\thepage} 
  
}

\theoremstyle{definition}
\newtheorem{theorem}{Theorem}
\newtheorem{definition}{Definition}
\newtheorem{ass}{Assumption}
\newtheorem{lem}{Lemma}
\newtheorem{rem}{Remark}
\newtheorem{exam}{Example}

\newcommand{\argmin}{\mathop{\rm arg\,min\,}\limits}

\title{Boosting-Based Sequential Meta-Tree Ensemble Construction for Improved Decision Trees}
\author{Ryota Maniwa, Naoki Ichijo, \\Yuta Nakahara, and Toshiyasu Matsushima}
\date{}

\begin{document}
  \pagestyle{fancy}
\setstretch{1.05}

\maketitle

\begin{abstract}
A decision tree is one of the most popular approaches in machine learning fields.
However, it suffers from the problem of overfitting caused by overly deepened trees. 
Then, a meta-tree is recently proposed.
It solves the problem of overfitting caused by overly deepened trees.
Moreover, the meta-tree guarantees statistical optimality based on Bayes decision theory. 
Therefore, the meta-tree is expected to perform better than the decision tree. 
In contrast to a single decision tree, it is known that ensembles of decision trees, which are typically constructed boosting algorithms, are more effective in improving predictive performance. 
Thus, it is expected that ensembles of meta-trees are more effective in improving predictive performance than a single meta-tree, and there are no previous studies that construct multiple meta-trees in boosting.
Therefore, in this study, we propose a method to construct multiple meta-trees using a boosting approach.
Through experiments with synthetic and benchmark datasets, we conduct a performance comparison between the proposed methods and the conventional methods using ensembles of decision trees. 
Furthermore, while ensembles of decision trees can cause overfitting as well as a single decision tree, experiments confirmed that ensembles of meta-trees can prevent overfitting due to the tree depth.
\end{abstract}
\newpage
\section{Introduction}
Prediction problem is a major research field in machine learning. The goal is to minimize the error between the predicted value and the unknown objective variable, given a set of training data and new explanatory variable data. The decision tree \cite{Breiman1984CARTCA} is a method in which trees are grown to divide the space of explanatory variables in order to minimize the splitting criterion, e.g., Gini Index and sample variance. Because of its expressivity and interpretability of such divisions, the decision tree is known as one of the most widely used prediction methods in machine learning.

However, a decision tree has the problem of overfitting. The depth of the tree is known to be crucial to its predictive performance. With trees that are too shallow, the space division will not be sufficient to express the data complexity. At the same time, trees that are too deep mistakenly capture non-existent dependency between data. Therefore, the predictive performance will be worse if the tree is too shallow or too deep. Pruning \cite{Breiman1984CARTCA} is often employed to reduce the depth of the single decision tree by cutting down its leaf nodes and choosing the single subtree of the original overgrown one. Another way to obtain a smaller tree is to add penalty terms to the splitting criterion in \cite{chen2016xgboost} and \cite{ke2017lightgbm}. The penalty term induces a smaller tree by controlling its depth or its number of leaf nodes.

To prevent overfitting, Suko~\cite{suko2003} and Dobashi~\cite{dobashi2021meta} proposed a tree structure called a meta-tree. The meta-tree is a set of trees. For any given tree, which is called a representative tree, the meta-tree is defined as the set of all the subtrees of the representative tree. Moreover, Suko~\cite{suko2003} and Dobashi~\cite{dobashi2021meta} proposed a prediction method using the meta-tree. In this method, all the subtrees in the meta-tree are used with appropriate weights, i.e., we can appropriately combine predictions of both shallow and deep trees. Therefore, this method prevents overfitting and provides a better prediction. Moreover, it is known that it takes only the same order of calculations as the method using the single representative tree when using all the subtrees in the meta-tree.

The prediction by the meta-tree is derived based on the statistical optimality in Bayes decision theory \cite{Berger:1327974}, treating the decision tree as a stochastic model. This tree-structured statistical model is called a model tree in \cite{suko2003} and \cite{dobashi2021meta}. They showed that the optimal prediction based on Bayes decision theory can be computed using a meta-tree.

In contrast to the prediction methods using a single decision tree, prediction methods using ensembles of decision trees are known to improve predictive performance. Thus, we can consider prediction methods using ensembles of meta-trees. Bagging and boosting are often used to construct ensembles of decision trees. 

In bagging, ensembles of decision trees are constructed independently by bootstrap sampling of the training data. Dobashi~\cite{dobashi2021meta} proposed a method to construct ensembles of meta-trees independently, as in a bagging approach. In boosting such as Gradient Boosting Decision Tree (GBDT) \cite{friedman2001greedy}, XGBoost \cite{chen2016xgboost}, and LightGBM \cite{ke2017lightgbm}, ensembles of decision trees are sequentially constructed using previously constructed decision trees. 

In this study, we propose a method to construct ensembles of meta-trees sequentially, as in the boosting approach. Boosting is more prone to overfitting than bagging. Therefore, it is necessary to determine the depth of the trees carefully. On the other hand, the advantage of the proposed method is that it can prevent overfitting even when increasing the depth of the meta-trees.

%
\newpage
\section{Preliminaries}
\label{sec:problem and bayes}

\subsection{Decision tree as a stochastic model}
\label{sec:model_tree}
In this section, we define the decision tree as a stochastic model.
Let the dimension of the continuous features and the dimension of the binary features be $p\in\{0\}\cup\mathbb{N}$ and $q\in\{0\}\cup\mathbb{N}$ respectively. Let the dimension of all the features represent $K=p+q$. Let $\bm{x}\coloneqq(x_1,\cdots,x_{p},x_{p+1},\cdots,x_{p+q})\in\mathcal{X}^K\coloneqq\mathbb{R}^p\times\{0,1\}^q$ represent the explanatory variable. Furthermore, we denote the objective variable as $y\in\mathcal{Y}$, and our method can be applied to a continuous set $\mathcal{Y}=\mathbb{R}$. Here, we first describe a triplet $(T,\bm{k},\boldsymbol{\theta})$ in Definition \ref{df:model_tree} and then define the probability structure $p(y|\bm{x},\boldsymbol{\theta},T,\bm{k})$ in Definition \ref{df:stochastic_model} as follows:

\begin{definition}\label{df:model_tree}
Let $T\in\mathcal{T}$ be a binary regular tree whose depth is equal to or smaller than $D_{\mathrm{max}}$. Let $\mathcal{S}_T$ represent the set of nodes for the tree $T$. $s\in\mathcal{S}_T$ represents the node in the tree $T$ and $s_\lambda$ represents the root node. Let $\mathcal{I}_T$ denote the set of internal nodes for $T$, and $\mathcal{L}_T$ denote the set of leaf nodes for $T$. The explanatory variable $x_{k_s}\in\mathcal{X} (k_s\in\{1,2,\cdots,K\})$ is assigned to the internal node $s\in\mathcal{I}_T$. If $x_{k_s}$ is a continuous variable, a threshold value $t_{k_s}$ is assigned to a node $s$. If $x_{k_s}$ is a binary variable, we put nothing in a threshold value $t_{k_s}$. We refer to $\bm{k}\coloneqq\left((k_s,t_{k_s})\right)_{s\in\mathcal{I}_T}\in\mathcal{K}$ as the features of the explanatory variable. Let $(T, \bm{k})$ denote model tree. Furthermore, a node $s\in\mathcal{S}_T$ has a parameter $\theta_s$ and we represent $\boldsymbol{\theta}\coloneqq(\theta_s)_{s\in\mathcal{S}_T}\in\boldsymbol{\Theta}$.
\end{definition}

\begin{figure}[t]
\begin{center}
\includegraphics[width=1\textwidth]{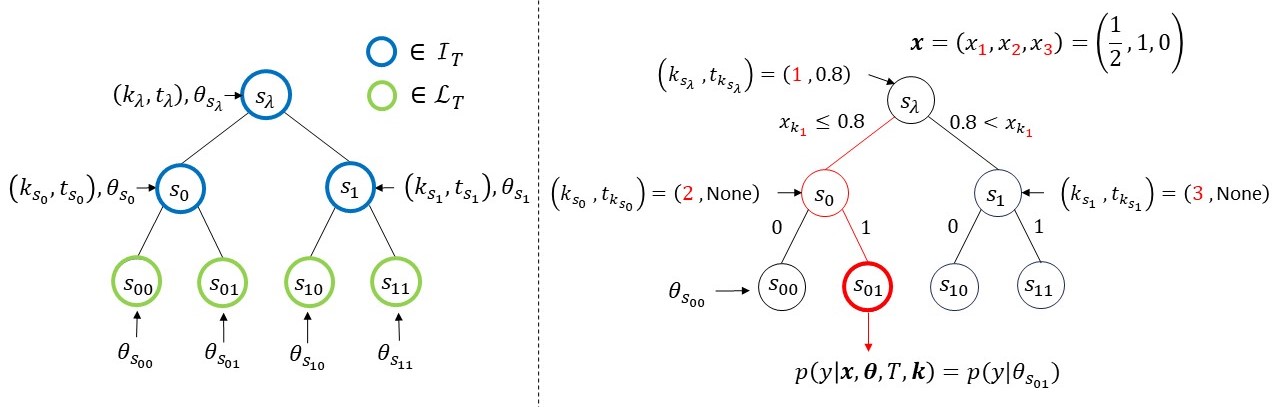}
\caption{The notations for the binary model tree (left) and an example of the model tree (right). The subscript of $\bm{x}$ (red) represents the feature $k_s$, and if $x_{k_s}$ is a continuous variable, it is divided by a threshold value $t_{k_s}$. If $x_{k_s}$ is a binary variable, it is divided by a binary value of 0 or 1. If $\bm{x}$ is assigned to the root node of the model tree (right), following the red path leads to the leaf node $s_{01}$. The output is generated from $p(y|\theta_{s_{01}})$.}\label{fig:model_tree}
\end{center}
\end{figure}

\begin{definition}
\label{df:stochastic_model}
The path from the root node to a leaf node in the model tree $(T,\bm{k})$ is uniquely determined by the explanatory variable $\bm{x}$. We denote the leaf node corresponding to $\bm{x}$ as $s_{T,\bm{k}}(\bm{x})\in\mathcal{L}_T$. Figure \ref{fig:model_tree} shows an example of that.
We define the stochastic model that represents the probability structure of $y$ given $\bm{x},\boldsymbol{\theta}, T,\bm{k}$ as follows:
\begin{align}\label{eq:stochastic_structure}
p(y|\bm{x},\boldsymbol{\theta},T,\bm{k})\coloneqq p(y|\theta_{s_{T,\bm{k}}(\bm{x})}).
\end{align}
Let $(T^*,\bm{k}^*)$ be the true model tree. We define the set of model trees that includes $(T^*,\bm{k}^*)$ as the set of candidate model trees. 
\end{definition}

Under the given $T,\bm{k}$, we assume a conjugate prior distribution for $\theta_s$.
\begin{ass}
\label{ass:distribution_theta}
  Let $p(\theta_s|T,\bm{k})$ be the conjugate prior distribution. For any node $s\in\mathcal{L}_T$, we assume the prior distribution on $\boldsymbol{\theta}$ as follows:
  \begin{align}
  \label{eq:distribution_theta}
    p(\boldsymbol{\theta}|T,\bm{k})=\prod_{s\in\mathcal{S}_T}p(\theta_s|T,\bm{k}).
  \end{align}
\end{ass}
\begin{exam}
The conjugate prior distribution of the normal distribution $\mathcal{N}(y|\mu_s,\tau_s^{-1})$ is the normal-gamma distribution $\mathcal{N}(\mu_s|m,(\kappa\tau_s)^{-1})\mathrm{Gam}(\tau_s|\alpha,\beta)$.
\end{exam}

Regarding a prior distribution of $T\in\mathcal{T}$, we impose the following assumption, which has been used in \cite{suko2003,dobashi2021meta,matsushima_bayes_coding,nakahara2022probability}. 
\begin{ass}
  \label{ass:distribution_T}
  Let $g_s\in[0,1]$ be a hyperparameter assigned to node $s\in\mathcal{S}_T$. In this study, we assume the prior distribution of $T\in\mathcal{T}$ as follows:
\begin{align}
\label{eq:distribution_T}
p(T)=\prod_{s\in\mathcal{I}_{T}}g_s\prod_{s'\in\mathcal{L}_{T}}(1-g_{s'}).
\end{align}
For $s\in\mathcal{L}_T$, we assume $g_s=0$. 
\end{ass}

\begin{figure}[t]
\begin{center}
\includegraphics[width=1\textwidth]{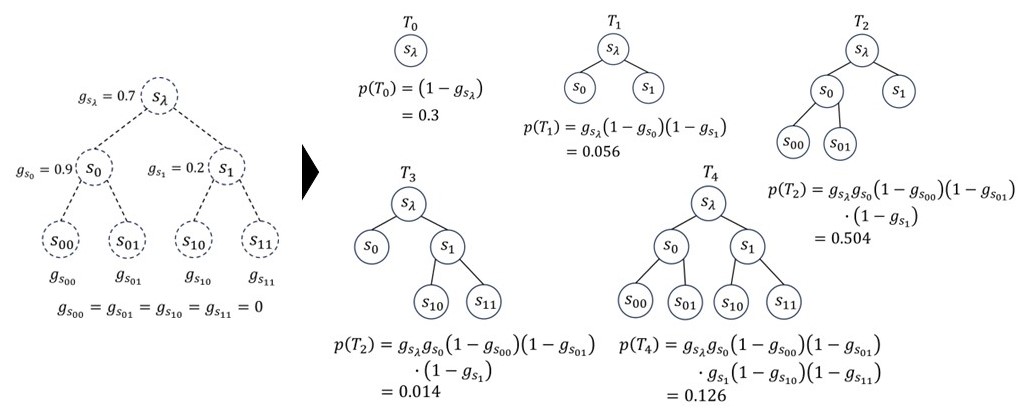}
\caption{An example of the prior distribution on $\{T_0,T_1,T_2,T_3,T_4\}\in\mathcal{T}$.}\label{fig:g}
\end{center}
\end{figure}

\subsection{Prediction with a single meta-tree}
\label{sec:k_known}
\subsubsection{Problem setup}
\label{problem}
In this study, we consider the regression problem. The purpose is to predict an unknown objective variable $y_{n+1}\in\mathcal{Y}$ corresponding to an explanatory variable $\bm{x}_{n+1}\in\mathcal{X}^K$ from given training data $(\bm{x}^n,y^n)\coloneqq\{(\bm{x}_i,y_i)\}_{i=1}^n$, where $n\in\mathbb{N}$ is the sample size. We assume that $y_i$ is generated independently according to the stochastic model $p(y|\bm{x},\boldsymbol{\theta}^*,T^*,\bm{k}^*)$ defined in Definition \ref{df:stochastic_model}. The true tree $T^*$ and the true parameters $\boldsymbol{\theta}^*$ are unknown, but we assume that the features $\bm{k}$, the set of candidate model trees $\mathcal{M}$, hyperparameters $(\theta_s)_{s\in\mathcal{S}_T}$ of $p(\boldsymbol{\theta})$ and $(g_s)_{s\in\mathcal{S}_T}$ of $p(T)$ are known.

Based on the above, we formulate the optimal prediction under Bayes decision theory \cite{Berger:1327974}  for any $\bm{k}$. In Bayes decision theory, mapping from the training data to the prediction value is called a ``decision function." Also, the Bayes risk function serves as an evaluation criterion for the decision function. In this study, we focus on regression problems. We define the decision function as $\delta:(\mathcal{X}^{K}\times\mathcal{Y})^n\times\mathcal{X}^K\times\mathcal{K}\rightarrow\mathcal{Y}$, and the Bayes risk function $\mathrm{BR}(\delta)$ based on the squared error loss $l(y_{n+1},\delta(\bm{x}^n,y^n,\bm{x}_{n+1},\bm{k}))$ is defined as follows.
\begin{align}\label{eq:Bayes_risk_function}
\mathrm{BR}(\delta)\coloneqq&\sum_{T\in\mathcal{T}}\int_{\boldsymbol{\theta}}p(\boldsymbol{\theta},T|\bm{k})\int_{\mathbb{R}^n}p(y^n|\bm{x}^n,\boldsymbol{\theta},T,\bm{k})\int_{\mathbb{R}}p(y_{n+1}|\bm{x}_{n+1},\boldsymbol{\theta},T,\bm{k})\nonumber\\
&\hspace{80pt}\times l(y_{n+1},\delta(\bm{x}^n,y^n,\bm{x}_{n+1},\bm{k}))dy_{n+1}dy^nd\boldsymbol{\theta}.
\end{align}
The optimal decision $\delta^*$ that minimizes the Bayes risk function $\mathrm{BR}(\delta)$ is given as Theorem \ref{theo:optimal_bayes_decision_function}. The proof of Theorem \ref{theo:optimal_bayes_decision_function} is given in Appendix A.
\begin{theorem}\label{theo:optimal_bayes_decision_function}
Under the assumption of squared error loss, the optimal decision function $\delta^*(\bm{x}^n,y^n,\bm{x}_{n+1},\bm{k})$ that minimizes $(\ref{eq:Bayes_risk_function})$ is given as follows:
\begin{align}
\label{eq:bayes_optimal_prediction}
&\delta^*(\bm{x}^n,y^n,\bm{x}_{n+1},\bm{k})\nonumber\\
&=\int_{\mathbb{R}}y_{n+1}\sum_{T\in\mathcal{T}}\int_{\boldsymbol{\Theta}} p(\boldsymbol{\theta},T|\bm{x}^n,y^n,\bm{k})p(y_{n+1}|\bm{x}_{n+1},\boldsymbol{\theta},T,\bm{k})d\boldsymbol{\theta}dy_{n+1}.
\end{align}
\end{theorem}
In this study, we call $\delta^*$ the Bayes optimal prediction. With the following procedures, we can obtain $\delta^*$ of (\ref{eq:bayes_optimal_prediction}).
\begin{enumerate}
  \item Weight the prediction from the stochastic model defined in (\ref{eq:stochastic_structure}) of Definition \ref{df:stochastic_model} with the posterior probabilities of $\boldsymbol{\theta}$ and $T$ given $\bm{x}^n,y^n$ and $\bm{k}$.
  \item Calculate expectation of $y_{n+1}$ for the values obtained in Step 1.
\end{enumerate}

\subsubsection{The Bayes optimal prediction using a meta-tree}
\label{sec:Bayes_optimal_prediction_meta-tree}
We rewrite (\ref{eq:bayes_optimal_prediction}) as follows.
\begin{align}
\label{eq:equation_q}
&q(y_{n+1}|\bm{x}_{n+1},\bm{x}^n,y^n,T,\bm{k})\nonumber\\
&\hspace{77pt}\coloneqq\int_{\boldsymbol{\Theta}}p(\boldsymbol{\theta}|\bm{x}^n,y^n,T,\bm{k})p(y_{n+1}|\bm{x}_{n+1},\boldsymbol{\theta},T,\bm{k})d\boldsymbol{\theta},\\
\label{eq:equation_tilde_q}
&\tilde{q}(y_{n+1}|\bm{x}_{n+1},\bm{x}^n,y^n,\bm{k})\nonumber\\
&\hspace{77pt}\coloneqq\sum_{T\in\mathcal{T}}p(T|\bm{x}^n,y^n,\bm{k})q(y_{n+1}|\bm{x}_{n+1},\bm{x}^n,y^n,T,\bm{k}),\\
\label{eq:bayes_optimal_prediction_q_reg}
&\delta^*(\bm{x}^n,y^n,\bm{x}_{n+1},\bm{k})=\int_{\mathbb{R}}y_{n+1}\tilde{q}(y_{n+1}|\bm{x}_{n+1},\bm{x}^n,y^n,\bm{k})dy_{n+1}.
\end{align}

\begin{rem}
\label{rem:q}
It is possible to solve (\ref{eq:equation_q}) analytically from Assumption \ref{ass:distribution_theta}.
\end{rem}
\begin{rem}\label{rem:tilde_q}
In calculating (\ref{eq:equation_tilde_q}), the issue arises where the computational complexity increases exponentially as the depth of the trees increases.
\end{rem}
It is known that (\ref{eq:equation_tilde_q}) can be analytically or efficiently solved by \cite{suko2003,dobashi2021meta} using a meta-tree. 

\begin{figure}[t]
\begin{center}
\includegraphics[width=1\textwidth]{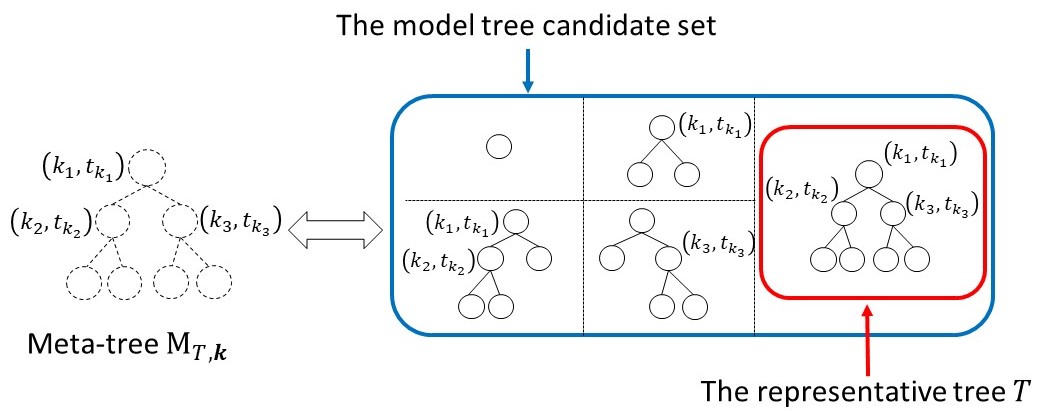}
\caption{An example of the meta-tree $\mathrm{M}_{T,\bm{k}}(\bm{k}=((k_1,t_{k_1}),(k_2,t_{k_2}),(k_3,t_{k_3})))$. The meta-tree represents the model tree candidate set enclosed in blue, and the largest model tree within the set (enclosed in red) is called the representative tree.}\label{fig:metatree}
\end{center}
\end{figure}

\begin{definition}
  We define a meta-tree $\mathrm{M}_{T,\bm{k}}$ as follows:
\begin{align}
\mathrm{M}_{T,\bm{k}}\coloneqq\{(T',\bm{k}')\in\mathcal{T}\times\mathcal{K} \mid\bm{k}'=\bm{k},T' \text{ is a subtree of } T\}.
\end{align}
We call $T$ a representative tree. Let $\mathcal{T}_{\mathrm{M}_{T,\bm{k}}}$ denote a set of $T$ which is represented by meta-tree $\mathrm{M}_{T,\bm{k}}$. We rewrite the model tree candidate set as $\mathcal{M}=\mathcal{T}_{\mathrm{M}_{T,\bm{k}}}\times\{\bm{k}\}$.
\end{definition}
In other words, the meta-tree is a set of multiple model trees (the model tree candidate set) subtrees of a single representative tree. Figure \ref{fig:metatree} shows it in detail.

By using a meta-tree, we represent (\ref{eq:equation_tilde_q}) as follows.
\begin{align}
&\tilde{q}(y_{n+1}|\bm{x}_{n+1},\bm{x}^n,y^n,\bm{k})\nonumber\\
\label{eq:equation_tilde_q_metatree}
&=\sum_{T\in\mathcal{T}_{\mathrm{M}_{T,\bm{k}}}}p(T|\bm{x}^n,y^n,\bm{k})q(y_{n+1}|\bm{x}_{n+1},\bm{x}^n,y^n,T,\bm{k})\\
&\eqqcolon \tilde{q}_{\mathrm{M}_{T,\bm{k}}}(y_{n+1}|\bm{x}_{n+1},\bm{x}^n,y^n).
\end{align}
We can calculate (\ref{eq:equation_tilde_q_metatree}) from Assumption \ref{ass:distribution_T}. The detailed solution method is given in Appendix B.
Furthermore, we represent the Bayes optimal prediction $\delta^*$ by the function $f$ defined in Definition \ref{df:function_f}.
\begin{definition}
\label{df:function_f}
we define the function $f$ that outputs the prediction corresponding to $\bm{x}_{n+1}$ from the meta-tree $\mathrm{M}_{T,\bm{k}}$ as follows:
\begin{align}
\label{eq:metatree_function}
f(\bm{x}_{n+1}|\mathrm{M}_{T,\bm{k}})&\coloneqq \int_{\mathbb{R}}y_{n+1}\tilde{q}_{\mathrm{M}_{T,\bm{k}}}(y_{n+1}|\bm{x}_{n+1},\bm{x}^n,y^n)dy_{n+1}\nonumber\\
&=\delta^*(\bm{x}^n,y^n,\bm{x}_{n+1},\bm{k}).
\end{align}  
\end{definition}

\begin{rem}
\label{rem:metatree}
Suko~\cite{suko2003} and Dobashi~\cite{dobashi2021meta} described that a meta-tree has the following properties.
\begin{itemize}
  \item If the true model tree is included in a meta-tree, the Bayes optimal prediction can be calculated by (\ref{eq:metatree_function}).
  \item By using a meta-tree, the computational cost of (\ref{eq:metatree_function}) can be reduced to only $\mathcal{O}(nD_{\mathrm{max}})$.
\end{itemize}
\end{rem}

\begin{rem}
Here, we compare the prediction based on the decision tree and the meta-tree. The predictive performance of a decision tree improves when configuring its depth to align with the characteristics of the true model tree. However, it is difficult to set the appropriate tree depth. 
If the depth of the decision tree is too shallow, the performance is worse due to poor data splitting, and if the depth of the decision tree is too deep, the performance is worse due to overfitting.
On the other hand, the prediction method using a meta-tree guarantees the Bayes optimality if the true model tree is included in a meta-tree as described in Remark \ref{rem:metatree}. Therefore, the depth of the meta-tree needs to be set deeper, which is an outstanding characteristic of the meta-tree.
\end{rem}

\subsection{Prediction with ensembles of $B$ meta-trees}
In Chapter \ref{sec:k_known}, we described that the prediction with a single meta-tree under given $\bm{k}$ guarantees the Bayes optimality. In this chapter, we assume that $\bm{k}$ is unknown, and we consider predicting with ensembles meta-trees.

\subsubsection{Problem setup}
The true model tree $(T^*,\bm{k}^*)$ and the true parameters $\boldsymbol{\theta}^*$ are unknown, but we assume that the set of candidate model trees $\mathcal{M}$, hyperparameters $(\theta_s)_{s\in\mathcal{S}_T}$ of $p(\boldsymbol{\theta})$ and $(g_s)_{s\in\mathcal{S}_T}$ of $p(T)$ are known. 

We consider a method to construct ensembles of meta-trees and predict using these meta-trees. 
We define $\mathcal{K}_B\coloneqq\{\bm{k}_1,\cdots,\bm{k}_B\}$ for $B$ meta-trees $\mathrm{M}_{T,\bm{k}_1},\cdots,\mathrm{M}_{T,\bm{k}_B}$. Furthermore, for $j=1,\cdots,B$, we define $f_j$ as follows.
\begin{align}
\label{eq:metatree_function_j}
f_j(\bm{x}_{n+1}|\mathrm{M}_{T,\bm{k}_j})&\coloneqq \int_{\mathbb{R}}y_{n+1}\tilde{q}_{\mathrm{M}_{T,\bm{k}_j}}(y_{n+1}|\bm{x}_{n+1},\bm{x}^n,y^n)dy_{n+1}.
\end{align}

We need to determine the features of explanatory variable $\bm{k}_1,\cdots,\bm{k}_B$, which is assigned to the $B$ meta-trees. And, for $j=1,\cdots,B$, let $w_j\in\mathbb{R}$ be the weight of $f_j$. 
In this case, we define $F_B$ as follows.
\begin{definition}
We define $F_B$ which is the prediction with $B$ meta-trees as follows:
\begin{align}
\label{eq:bayes_optimal_prediction_q_reg_approximation}
F_B(\bm{x}_{n+1})\coloneqq\sum_{j=1}^Bw_jf_j(\bm{x}_{n+1}|\mathrm{M}_{T,\bm{k}_j}).
\end{align}  
\end{definition}
Dobashi~\cite{dobashi2021meta} proposed Meta-Tree Random Forest (MTRF) as a method to construct ensembles of $B$ meta-trees. 
MTRF independently constructs $B$ meta-trees similar to Random Forest \cite{breiman2001random}. Dobashi~\cite{dobashi2021meta} calculated (\ref{eq:bayes_optimal_prediction_q_reg_approximation}) by regarding $w_j$ as the posterior distribution of $\bm{k}\in\mathcal{K}_B$. This detail is described in Chapter \ref{sec:distribution_k}.
\newpage
\section{Sequential construction of meta-trees}
\label{sec:proposed_methods}
In this chapter, we describe the main results of our study. To compute (\ref{eq:bayes_optimal_prediction_q_reg_approximation}), we propose a method to construct ensembles of $B$ meta-trees using a boosting approach. This approach involves using the trees already obtained to construct a new tree.

\subsection{Evaluation function for meta-trees}
Let the evaluation function $\mathrm{E}$ of the prediction $F_B$ represent the mean squared error described by the following equation.
\begin{align}
\mathrm{E}(F_B) &\coloneqq \sum_{i=1}^n \mathrm{L}
\left(y_i,F_B(\bm{x}_i)\right) \\
\label{eq:evaluation}
&\coloneqq \sum_{i=1}^n\left(y_i - F_B(\bm{x}_i)\right)^2.
\end{align}
It is difficult to find $B$ meta-trees $(\mathrm{M}_{T,\bm{k}_1},\cdots,\mathrm{M}_{T,\bm{k}_B})$ simultaneously that minimize (\ref{eq:evaluation}). Therefore, for $b=1,\cdots,B$, we define as follows.
\begin{align}\label{eq:F_b}
F_b(\bm{x}_i)\coloneqq \sum_{j=1}^bw'_jf_j(\bm{x}_i|\mathrm{M}_{T,\bm{k}_j}).
\end{align}
We denote the learning weights as $\{w'_j\}_{j=1}^b$, which use in (\ref{eq:F_b}), and denote the predicting weights as $\{w_j\}_{j=1}^b$, which use in (\ref{eq:bayes_optimal_prediction_q_reg_approximation}). 
Furthermore, with $\{\mathrm{M}_{T,\bm{k}_j}\}_{j=1}^{b-1}$ already determined, we consider finding $\mathrm{M}_{T,\bm{k}_b}$ such that $\mathrm{E}(F_b)$ is minimized. To find it, we reduce the gradient of $\mathrm{E}(F_b)$. For the mean squared error such as (\ref{eq:evaluation}), the gradient becomes the residual $r_i\coloneqq y_i - F_{b-1}(\bm{x}_i)$. Using CART \cite{Breiman1984CARTCA}, a representative tree $(T,\bm{k}_b)$ is constructed by splitting the data so that the sample variance of the residuals $\{r_i\}_{i=1}^n$ is minimized. Moreover, the meta-tree $\mathrm{M}_{T,\bm{k}_b}$ is obtained from $(T,\bm{k}_b)$. We summarize the proposed method in Algorithm \ref{alg:sequential_algorithm}. The two functions used in Algorithm \ref{alg:sequential_algorithm} are described as follows.
\begin{itemize}
  \item $\mathrm{BuildTree}(\mathcal{D}_b,D_\mathrm{max})$: A function that takes input of training data $\mathcal{D}_b$ and maximum depth $D_\mathrm{max}$, and outputs a representative tree $(T,\bm{k}_b)$ using CART algorithm.
  \item $\mathrm{BuildMetaTree}((T,\bm{k}_b))$: A function to convert a representative tree to a meta-tree.
\end{itemize}

\section{Prediction of $F_b(\bm{x})$ and $F_B(\bm{x})$}
\subsection{Model based on GBDT}
We consider a method for prediction of $F_b(\bm{x})$ used in conventional methods. Similar to the GBDT \cite{friedman2001greedy}, the output of the $b$-th meta-tree is learned to be the residual between $F_{b-1}$ and $y$. In this case, the learning and predicting weights are denoted as $\{w'_j=1\}_{j=1}^b$ and $\{w_j=1\}_{j=1}^B$.

\subsection{Model with weights as probability distribution}
We can also consider meta-trees where each one predicts $y$. In this case, some probability distribution is available for the training and predicting weights.
\subsubsection{A uniform distribution}
Random Forest \cite{breiman2001random} calculates the average of the prediction from the constructed trees. Therefore, it can be regarded as using uniform probabilities as the weights. The learning and predicting weights are denoted as $\{w'_j=1/b\}_{j=1}^b$ and $\{w_j = 1/B\}_{j=1}^B$.

\subsubsection{A posterior distribution of $\bm{k}$}\label{sec:distribution_k}
The expected value of (\ref{eq:metatree_function}) by the posterior distribution of $\bm{k}\in\mathcal{K}$ results in the Bayes optimal prediction under the unknown $\bm{k}$.
Therefore, we consider the posterior of $\bm{k}$ as both the learning and the predicting weights. We denote the posterior distribution of $\bm{k}\in\mathcal{K}_B$ as $p_{\mathcal{K}_B}(\bm{k}|\bm{x}^n,y^n)$. Dobashi~\cite{dobashi2021meta} calculated (\ref{eq:bayes_optimal_prediction_q_reg_approximation}) by regarding $w_j$ as $p_{\mathcal{K}_B}(\bm{k}_j|\bm{x}^n,y^n)$. By assuming that the prior distribution of $\bm{k}\in\mathcal{K}$ be a uniform distribution over $\mathcal{K}_B$, the posterior distribution of $\bm{k}$ is obtained. Dobashi~\cite{dobashi2021meta} proved that Lemma \ref{lem:distribution_k} show a posterior probability distribution of $\bm{k}\in\mathcal{K}_B$. The proof of Lemma \ref{lem:distribution_k} is given in Appendix C.

\begin{lem}\label{lem:distribution_k}
By using $\tilde{q}_{\mathrm{M}_{T,\bm{k}}}(y_{n+1}|\bm{x}_{n+1},\bm{x}^n,y^n)$ as in $(\ref{eq:equation_tilde_q_metatree})$, a posterior probability distribution of $\bm{k}\in\mathcal{K}_B$ is expressed as follows:
\begin{align}
p_{\mathcal{K}_B}(\bm{k}|\bm{x}^n,y^n)\propto \prod_{i=1}^n\tilde{q}_{\mathrm{M}_{T,\bm{k}}}(y_i|\bm{x}_i,\bm{x}^{i-1},y^{i-1}).
\end{align}
\end{lem}

In proposed method, by assuming that the prior distribution of $\bm{k}\in\mathcal{K}$ be the uniform distribution over $\mathcal{K}_{b-1}$, the learning weights are obtained as $\{w'_j=p_{\mathcal{K}_{b-1}}(\bm{k}_j|\bm{x}^n,y^n)\}_{j=1}^{b-1}$. Similarly, the predicting weights are represented as $\{w_j=p_{\mathcal{K}_B}(\bm{k}_j|\bm{x}^n,y^n)\}_{j=1}^B$.

\subsection{Regularization by shrinkage}
In boosting, predicted values are scaled by a learning rate, slowing the learning speed by amplifying residuals, thereby preventing overfitting. Therefore, incorporating the learning rate $\gamma$ where $0<\gamma\leq1$ into the proposed method has the potential for improved performance.

\begin{algorithm}[tb]
\caption{Meta-trees construction Algorithm}
\label{alg:sequential_algorithm}
\begin{algorithmic}[1] 
\Require $\{(\bm{x}_i,y_i)\}_{i=1}^n$: the training data, $B$: the number of trees, $D_{\mathrm{max}}$: the maximum of the depth, $\gamma$: the learning rate
\Ensure $F_B(\bm{x})$: the prediction with $B$ meta-trees
\State $\mathcal{M}=\{\phi\},F_0(\bm{x}_i) = 0\hspace{10pt}(i=1,\cdots,n)$
\If{Model based GBDT}
\State $F_0(\bm{x})=\argmin_{\alpha\in\mathbb{R}}\sum_{i=1}^n\mathrm{L}(y_i,\alpha)$
\EndIf
\For{$b=1,\cdots,B$}
\If{$b\not=1$}
\State Select $\{w'_1,\cdots,w'_{b-1}\}$ \Comment{learning weights}
\State $F_{b-1}=\sum_{j\in\{1,\cdots,b-1\}}w'_jf_j\hspace{10pt}(i=1,\cdots,n)$
\EndIf
\State $r_{i}=y_i-\gamma F_{b-1}\hspace{10pt}(i=1,\cdots,n)$ 
\State $\mathcal{D}_b=\{(\bm{x}_i,r_i)\}_{i=1}^n$
\State $(T,\bm{k}_b)\gets \mathrm{BuildTree}(\mathcal{D}_b,D_\mathrm{max})$
\State $\mathrm{M}_{T,\bm{k}_b}\gets \mathrm{BuildMetaTree}((T,\bm{k}_b))$ 
\State $\mathcal{M}\gets\mathcal{M}\cup\{\mathrm{M}_{T,\bm{k}_b}\}$
\EndFor
\State Select $\{w_1,\cdots,w_B\}$ \Comment{predicting weights}
\State $F_B=\sum_{j\in\{1,\cdots,B\}}w_jf_j$
\State {\bf Return} $F_B(\bm{x})$
\end{algorithmic}
\end{algorithm}

\section{Experiments}
\label{sec:experiment_section}
In this chapter, we compare the performance of the proposed methods with conventional methods using synthetic and benchmark datasets. The following summarizes the methods to be compared.
\begin{itemize}
  \item {\bf Proposed methods}
  \begin{itemize}
    \item {\bf MT\_gbdt:} The model based on GBDT. It multiplies the learning rate $\gamma=0.1$ by the residuals.
    \item {\bf MT\_uni-uni:} The model with weights as a probability distribution. Both the learning weights and the prediction weights are uniform probabilities. The learning rate is $\gamma=1.0$.
    \item {\bf MT\_uni-pos:} The model with weights as a probability distribution. The learning weights are the uniform probabilities, and the prediction weights are the posterior probabilities of $\bm{k}$. The learning rate is $\gamma=1.0$.
    \item {\bf MT\_pos-pos:} The model with weights as a probability distribution. Both the learning weights and the prediction weights are the posterior probability of $\bm{k}$. The learning rate is $\gamma=1.0$.
  \end{itemize}
  \item {\bf Conventional methods}
  \begin{itemize}
    \item {\bf GBDT:} the Gradient Boosting Decision Tree \cite{friedman2001greedy}. $D_{\mathrm{max}}$ and $B$ are the same as in the proposed methods. The criterion is "squared\_error," and the learning rate is $\gamma=0.1$. Whereas the other parameters use default values.
    \item {\bf LightGBM:} the LightGBM \cite{ke2017lightgbm}. $D_{\mathrm{max}}$ and $B$ are the same as in the proposed method, whereas the other parameters use default values.
  \end{itemize}
\end{itemize}

\subsection{Experiment 1}
\label{sec:exp_1}
\begin{figure}[t]
\centering
\includegraphics[width=0.5\textwidth]{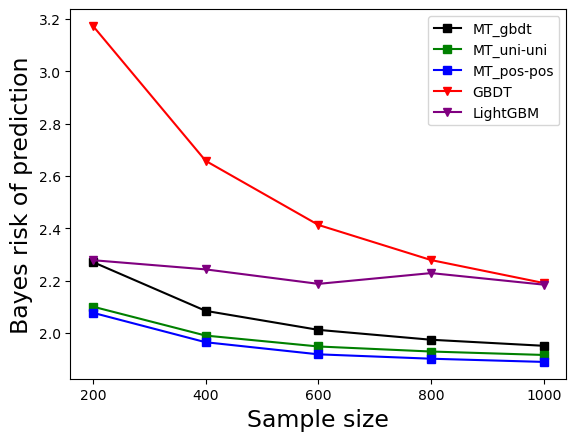}
\caption{The result of Experiment 1}\label{fig:bayesrisk_train_data}
\end{figure}

{\bf Purpose:} The proposed methods are used for the Bayes optimal prediction, which minimizes the Bayes risk. Therefore, the Bayes risk of the proposed methods is expected to be smaller than that of the conventional method. In this experiment, we confirm the Bayes optimality of the proposed methods using synthetic data generated from the true model tree. We compare the proposed methods (MT\_gbdt, MT\_uni-uni, and MT\_pos-pos) and the conventional methods (GBDT and LightGBM).\\\\
{\bf Conditions:} We generate true model trees by randomly assigning $\bm{k}^*=(k_s)_{s\in\mathcal{S}_{T^*}}$ of $K=10 (p=0,q=10)$ to the nodes of binary tree structures $T^*$ with a maximum depth of $D_{\mathrm{max}}^*=3$, $g_s=0.9$, and we determine parameters $\boldsymbol{\theta}^*$ according to the normal-gamma distribution $\mathcal{N}(\mu_s|0,(2\tau_s)^{-1})\mathrm{Gam}(\tau_s|2,2)$. The true model tree is generated 100 times. We generate $1000$ train data and $250$ test data $10$ times from each true model tree. Then, we perform each method using training data $n=200,400,600,800,1000$ with the following settings. $D_{\mathrm{max}}=5$, $B=100$, $g_s=0.6$. Afterward, we calculate each method's mean squared error (MSE) using test data. \\\\
{\bf Result:} Figure \ref{fig:bayesrisk_train_data} shows the approximated Bayes risk of the prediction. The Bayes risk of the proposed methods was smaller than that of the conventional methods (GBDT and LightGBM). Because the constructed ensembles of $B$ meta-trees include the subtree closely resembling the true model tree, the Bayes risk of the method using the posterior probabilities of $\bm{k}$ (MT\_pos-pos) was the smallest.

\subsection{Experiment 2}
\begin{figure}[t]
\begin{minipage}{0.5\hsize}
\centering
\includegraphics[width=1\textwidth]{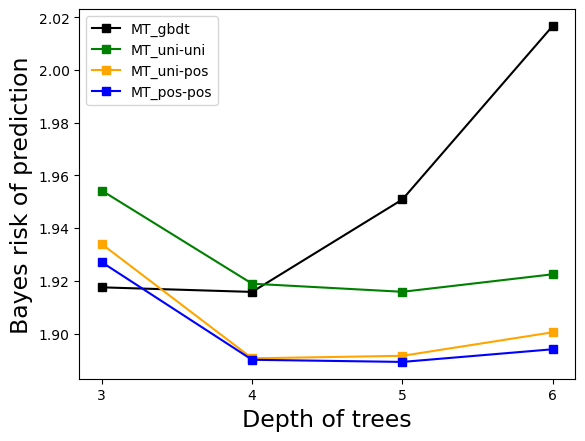}
\subcaption{$D_{\mathrm{max}}^*=3$}\label{fig:bayesrisk_depth3}
\end{minipage}
\begin{minipage}{0.5\hsize}
\centering
\includegraphics[width=1\textwidth]{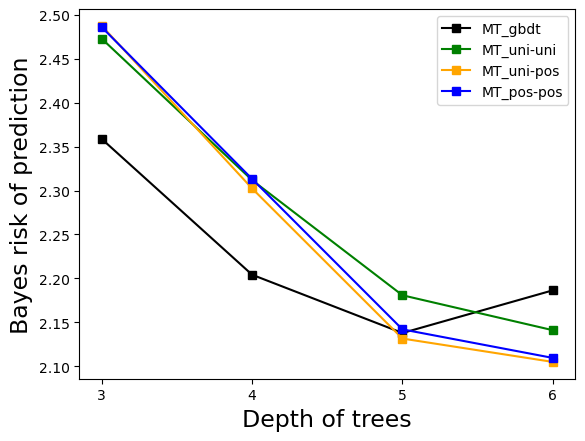}
\subcaption{$D_\mathrm{max}^*=5$}\label{fig:bayesrisk_depth5}
\end{minipage}
\begin{minipage}{0.5\hsize}
\centering
\includegraphics[width=1\textwidth]{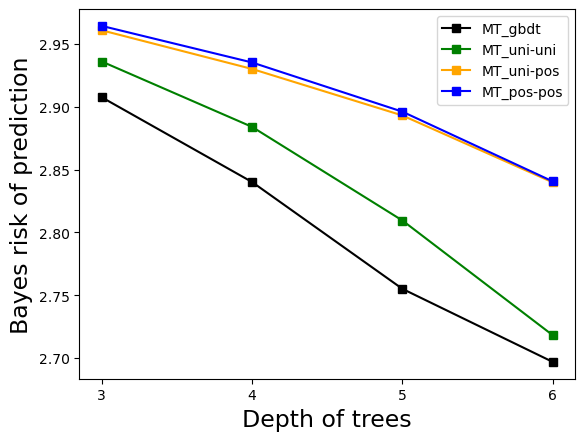}
\subcaption{$D_\mathrm{max}^*=7$}\label{fig:bayesrisk_depth7}
\end{minipage}
\caption{The result of Experiment 2. The maximum depth of the true model tree is 3,5,7, and the result of increasing the depth of the meta-trees to 3,4,5,6 is shown in (a), (b), and (c).}\label{fig:bayesrisk_depth}
\end{figure}

{\bf Purpose:} In Experiment 2, We confirm the influence between the depth of a true model tree and the depth of $B$ meta-trees. In addition, we compare the learning and prediction weights for various maximum depths of the true model tree. We compare the proposed methods (MT\_gbdt, MT\_uni-uni, MT\_uni-pos, and MT\_pos-pos).\\\\
{\bf Conditions:} We set the parameters of the true model tree to $D_{\mathrm{max}}^*=3,5,7$, $g_s=0.9$, and we determine parameters $\boldsymbol{\theta}^*$ according to the normal-gamma distribution $\mathcal{N}(\mu_s|0,(2\tau_s)^{-1})\mathrm{Gam}(\tau_s|2,2)$, and generate data using the same procedure as in Experiment 1. We perform each method using training data $n=1000$ with the following settings. We assume $D_{\mathrm{max}}=3,4,5,6$, $B=100$, $g_s=0.6$. Afterward, we calculate the mean squared error of each method using test data. \\\\
{\bf Result:} The result is shown in Figure \ref{fig:bayesrisk_depth}. In the case of (a) in Figure \ref{fig:bayesrisk_depth}, as the depth of the meta-trees increased, the Bayes risk of the methods using the posterior probabilities of $\bm{k}$ was smaller than that of the other methods. This is because the true model tree can be included inside the meta trees. Additionally, whether the learning weights are based on the posterior probabilities of $\bm{k}$ or the uniform probabilities, there is a tendency for little difference in the Bayes risk. Furthermore, there is a tendency for less overfitting to occur even when increasing the depth in cases MT\_pos-pos, MT\_uni-pos, and MT\_uni-uni. As shown in (b) and (c) in Figure \ref{fig:bayesrisk_depth}, when the depth of the meta-trees was shallower than the depth of the true model tree, the predictive performance of the methods using posterior probabilities was worse. This is because the true model tree can not be included inside the meta trees. In this case, the Bayes risk of MT\_gbdt and MT\_uni-uni were smaller than the other methods. Therefore, MT\_gbdt and MT\_uni-uni can be inferred to be effective methods when the true model tree cannot be included in the meta-trees.

\subsection{Experiment 3}
{\bf Purpose:} In Experiment 3, we confirm the predictive performance of the proposed methods (MT\_gbdt, MT\_uni-uni, and MT\_pos-pos) on multiple benchmark datasets for regression. Moreover, we confirm that the overfitting can be prevented when the tree depth is increased compared to the conventional methods (GBDT and LightGBM).\\\\
{\bf Conditions:} We summarize the details of each dataset in Table\ref{tab:dataset}. Abalone, Liver, and Student datasets are collected from UCI Machine Learning Repository \cite{Newman+Hettich+Blake+Merz:1998}. Diabetes dataset is collected from scikit-learn 1.3.2 \cite{pedregosa2018scikitlearn}. Cps datasets are collected from \cite{berndt1991practice}. Ozone dataset is from \cite{ozone}. 

We preprocess all datasets in the following steps.
\begin{enumerate}
  \item Remove missing values included in the dataset.
  \item Standardize the continuous explanatory and objective variables to align the scales.
  \item Apply label encoding for discrete variables with ordinal scales and one-hot vector encoding for discrete variables with nominal scales.
\end{enumerate}

\begin{table}[t]
    \centering
    \caption{List of datasets. For each dataset, the number of samples and attributes is listed. $q^*$ denotes the number of discrete variables before one-hot vector encoding.}
    \label{tab:dataset}
    \begin{tabular}{l|rrr}
        \toprule
        {\bf Dataset} & $n$ & $p$ & $q^*$ \\
        \midrule
        Abalone & $4177$ & $7$ & $1$\\
        Cps & $534$ & $3$ & $7$\\
        Diabetes & $442$ & $9$ & $1$\\
        Liver & $345$ & $5$ & $0$\\
        Ozone & $330$ & $8$ & $0$\\
        Student & $395$ & $13$ & $17$\\
        \bottomrule
    \end{tabular}
\end{table}
We set the parameters as follows. $D_{\mathrm{max}}=4,8$, $B=100$, $g_s=0.6$. We perform the 5-fold cross-validation tree times and output the average of the MSE.\\\\
{\bf Result:} The result is shown in Table \ref{tab:result_benchmark}. In most cases, the proposed methods were more accurate than the conventional method for benchmark datasets. Among the proposed methods, MT\_pos-pos did not perform well, while the MSE for MT\_gbdt and MT\_uni-uni were better. From the results of Experiment 2, it can be inferred that these datasets have difficulty representing the true model tree. Furthermore, when the tree depth was increased, the proposed method prevented overfitting compared to the conventional methods such as GBDT and LightGBM.

\begin{table}[ht]
    \centering
     \caption{The result of Experiment 3. The result with the smallest MSE in each dataset is in bold gray, and the result with the second smallest MSE is in gray.}
    \label{tab:result_benchmark}
    \small
    \makebox[1 \textwidth][c]{
    \begin{tabular}{l|rr|rr|rr|rr|rr}
        \toprule
        {\bf Methods} & \multicolumn{2}{c|}{\bf MT\_gbdt} & \multicolumn{2}{c|}{\bf MT\_uni-uni} & \multicolumn{2}{c|}{\bf MT\_pos-pos} & \multicolumn{2}{c|}{\bf GBDT} & \multicolumn{2}{c}{\bf LightGBM}\\
        {\bf Depth} & $4$ & $8$ & $4$ & $8$ & $4$ & $8$ & $4$ & $8$ & $4$ & $8$\\
        \midrule
      Abalone & $\cellcolor[gray]{0.85}{0.452}$ & $0.454$ & $0.506$ & $0.461$ & $0.542$ & $0.514$ & $0.453$ & $0.500$ & $\cellcolor[gray]{0.85}{\bf 0.446}$ & $0.454$\\
      Cps & $0.758$ & $0.779$ & $\cellcolor[gray]{0.85}{0.754}$ & $\cellcolor[gray]{0.85}{0.754}$ & $0.828$ & $0.828$ & $0.865$ & $1.07$ & $\cellcolor[gray]{0.85}{\bf 0.749}$ & $0.786$\\
      Diabetes & $\cellcolor[gray]{0.85}{\bf 0.565}$ & $0.577$ & $0.582$ & $\cellcolor[gray]{0.85}{0.573}$ & $0.681$ & $0.682$ & $0.617$ & $0.751$ & $0.591$ & $0.598$\\
      Liver & $\cellcolor[gray]{0.85}{0.839}$ & $0.850$ & $\cellcolor[gray]{0.85}{\bf 0.838}$ & $0.846$ & $0.903$ & $0.913$ & $0.937$ & $1.23$ & $0.905$ & $0.934$\\
      Ozone & $\cellcolor[gray]{0.85}{0.285}$ & $\cellcolor[gray]{0.85}{\bf 0.284}$ & $0.293$ & $0.289$ & $0.347$ & $0.341$ & $0.311$ & $0.383$ & $0.297$ & $0.309$\\
      Student & $\cellcolor[gray]{0.85}{\bf 0.812}$ & $\cellcolor[gray]{0.85}{0.825}$ & $0.842$ & $0.827$ & $0.918$ & $0.927$ & $0.892$ & $1.07$ & $0.862$ & $0.891$\\
      \bottomrule
    \end{tabular}
    }
\end{table}

\newpage
\section{Conclusion}
In this study, we proposed a method to construct ensembles of meta-trees as in a boosting approach. Although it is difficult to set the appropriate tree depth for conventional methods that construct ensembles of decision trees, the proposed method prevents overfitting even when the depth is set deep enough. The properties of the proposed method were confirmed using synthetic datasets assuming a true model tree and some well-known benchmark datasets.

\newpage
\thispagestyle{myheadings}
\appendix
\section{Proof of Theorem \ref{theo:optimal_bayes_decision_function}}
We prove Theorem \ref{theo:optimal_bayes_decision_function}. We solve the Bayes optimal prediction $\delta^*$ that minimizes $(\ref{eq_app:Bayes_risk_function})$.
\begin{proof}
\begin{align}\label{eq_app:Bayes_risk_function}
\mathrm{BR}(\delta)=\sum_{T\in\mathcal{T}}\int_{\boldsymbol{\theta}}p(\boldsymbol{\theta},T|\bm{k})\int_{\mathbb{R}^n}p(y^n|\bm{x}^n,&\boldsymbol{\theta},T,\bm{k})\int_{\mathbb{R}}p(y_{n+1}|\bm{x}_{n+1},\boldsymbol{\theta},T,\bm{k})\nonumber\\
&\times l(y_{n+1},\delta)dy_{n+1}dy^nd\boldsymbol{\theta}.
\end{align}
$(\ref{eq_app:Bayes_risk_function})$ can be altered as follows.
\begin{align}
\mathrm{BR}(\delta)
&=\int_{\mathbb{R}^n}\int_\mathbb{R}(y_{n+1}-\delta)^2\nonumber\\
&\hspace{30pt}\times\underset{\coloneqq p(y_{n+1}|\bm{x}_{n+1},\bm{x}^n,y^n)}{\underline{\left(\sum_{\mathcal{T}}\int_{\boldsymbol{\theta}}p(\boldsymbol{\theta},T|\bm{x}^n,y^n,\bm{k})p(y_{n+1}|\bm{x}_{n+1},\boldsymbol{\theta},T,\bm{k})d\boldsymbol{\theta}\right)}}\nonumber\\
&\hspace{200pt}\times dy_{n+1}p(y^n)dy^n.\\
&=\int_{\mathbb{R}^n}\underset{\coloneqq f(\delta)}{\underline{\int_\mathbb{R}(y_{n+1}-\delta)^2p(y_{n+1}|\bm{x}_{n+1},\bm{x}^n,y^n)dy_{n+1}}}p(y^n)dy^n.
\end{align}
\begin{align}
f(\delta) &= \int_{\mathbb{R}}(y^2_{n+1}-2y_{n+1}\delta+\delta^2)p(y_{n+1}|\bm{x}_{n+1},\bm{x}^n,y^n)dy_{n+1}\\
&=\delta^2-2\delta\int_\mathbb{R}p(y_{n+1}|\bm{x}_{n+1},\bm{x}^n,y^n)dy_{n+1}\nonumber\\
& \hspace{100pt}+\int_{\mathbb{R}}y^2_{n+1}p(y_{n+1}|\bm{x}_{n+1},\bm{x}^n,y^n)dy_{n+1}\\
& = \left(\delta - \int_\mathbb{R}y_{n+1}p(y_{n+1}|\bm{x}_{n+1}\bm{x}^n,y^n)dy_{n+1}\right)^2\nonumber\\
&\hspace{30pt}-\int_\mathbb{R}y_{n+1}p(y_{n+1}|\bm{x}_{n+1}\bm{x}^n,y^n)dy_{n+1}\nonumber\\
&\hspace{60pt}+\int_\mathbb{R}y^2_{n+1}p(y_{n+1}|\bm{x}_{n+1}\bm{x}^n,y^n)dy_{n+1}.
\end{align}
Therefore, the Bayes optimal prediction that minimizes $\mathrm{BR}(\delta)$ is obtained as follows.
\begin{align}
\delta^*&= \argmin_{\delta}\mathrm{BR}(\delta)\\
&= \argmin_{\delta}f(\delta)\\
&= \int_\mathbb{R}y_{n+1}p(y_{n+1}|\bm{x}_{n+1}\bm{x}^n,y^n)dy_{n+1}
\end{align}
\end{proof}

\section{Calculation of equation $(\ref{eq:equation_tilde_q_metatree})$}

In this appendix, we describe a method to calculate $(\ref{eq:equation_tilde_q_metatree})$. First, Suko~\cite{suko2003} and Dobashi\cite{dobashi2021meta} denoted that $(\ref{eq:equation_q})$ has a important property as follows.
\begin{lem}\label{lem2.4.1} For any $T,T'\in\mathcal{T},\bm{x}^n\in\mathcal{X}^{Kn},y^n\in\mathcal{Y}^n,\bm{x}_{n+1}\in\mathcal{X}^K$, if $s(\bm{x}_{n+1})\in\mathcal{L}_T\cap\mathcal{L}_{T'}$, then
  \begin{align}\label{lem1}
  q(y_{n+1}|\bm{x}_{n+1},\bm{x}^n,y^n,T,\bm{k})=q(y_{n+1}|\bm{x}_{n+1},\bm{x}^n,y^n,T',\bm{k}).
  \end{align}
From this lemma, since both sides of (\ref{lem1}) are determined by $s(\bm{x}_{n+1})$ regardless of the tree structure, we denote them as $q_{s(\bm{x}_{n+1})}(y_{n+1}|\bm{x}_{n+1},\bm{x}^n,y^n,\bm{k})$.
\end{lem}

By using meta-tree $\mathrm{M}_{T,\bm{k}}$, Suko~\cite{suko2003} and Dobashi~\cite{dobashi2021meta} introduced a recursive function to calculate (\ref{eq:equation_tilde_q_metatree}).
\begin{definition}
  For any node $s\in\mathcal{S}_T$ corresponding to $\bm{x}_{n+1}$ in $\mathrm{M}_{T,\bm{k}}$, we define the recursive function $\tilde{q}_s(y_{n+1}|\bm{x}_{n+1},\bm{x}^n,y^n,\mathrm{M}_{T,\bm{k}})$ as follows. Here, $s_{\mathrm{child}}$ is a child node of $s\in\mathcal{S}_T$.
\begin{align}
&\tilde{q}_s(y_{n+1}|\bm{x}_{n+1},\bm{x}^n,y^n,\mathrm{M}_{T,\bm{k}})\nonumber\\
&:=
\begin{cases}
q_s(y_{n+1}|\bm{x}_{n+1},\bm{x}^n,y^n,\mathrm{M}_{T,\bm{k}})\hspace{85pt}(s\in\mathcal{L}_T)\\
(1-g_{s|\bm{x}^n,y^n})q_s(y_{n+1}|\bm{x}_{n+1},\bm{x}^n,y^n,\bm{k})\\
\hspace{10pt}+g_{s|\bm{x}^n,y^n}\tilde{q}_{s_\mathrm{child}}(y_{n+1}|\bm{x}_{n+1},\bm{x}^n,y^n,\mathrm{M}_{T,\bm{k}})\hspace{20pt}(s\in\mathcal{I}_T)\hspace{10pt}.
\end{cases}
\end{align}
$g_{s|\bm{x}^n,y^n}$ is calculated as follows:
\begin{align}
g_{s|\bm{x}^i,y^i}:=
\begin{cases}
g_s\hspace{160pt}(i=0)\\
\frac{g_{s|\bm{x}^{i-1},y^{i-1}}\tilde{q}_{s_{\mathrm{child}}}(y_i|\bm{x}_i,\bm{x}^{i-1},y^{i-1},\mathrm{M}_{T,\bm{k}})}{\tilde{q}_s(y_i|\bm{x}_i,\bm{x}^{i-1},y^{i-1},\mathrm{M}_{T,\bm{k}})}\hspace{15pt}(\mathrm{otherwise})\hspace{10pt}.
\end{cases}
\end{align}
\end{definition}
In this case, Suko~\cite{suko2003} and Dobashi~\cite{dobashi2021meta} proved that (\ref{eq:equation_tilde_q_metatree}) is calculated in Theorem \ref{tildeqsolve}.
\begin{theorem}\label{tildeqsolve}
  When considering $s_\lambda$ as root node of $\mathrm{M}_{T,\bm{k}}$,
\begin{align}
(\ref{eq:equation_tilde_q_metatree})=\tilde{q}_{s_{\lambda}}(y_{n+1}|\bm{x}_{n+1},\bm{x}^n,y^n,\mathrm{M}_{T,\bm{k}}).
\end{align}
\end{theorem}
Therefore, the Bayes optimal prediction with a single meta-tree $\mathrm{M}_{T,\bm{k}}$ can be expressed as follows:
\begin{align}
\label{bayesop3}
\delta^*(\bm{x}^n,y^n,\bm{x}_{n+1})=\int_{\mathbb{R}}y_{n+1}\tilde{q}_{s_\lambda}(y_{n+1}|\bm{x}_{n+1},\bm{x}^n,y^n,\mathrm{M}_{T,\bm{k}})dy_{n+1}.
\end{align}

\section{Proof of Lemma \ref{lem:distribution_k}}
\begin{proof}
We assume that the prior distribution of $\bm{k}\in\mathcal{K}$ be a uniform distribution on $\mathcal{K}_B$. In this case, Dobashi~\cite{dobashi2021meta} proved Lemma\ref{lem:distribution_k} as follows:
\begin{align}
p(\bm{k}|\bm{x}^n,y^n)&\propto p(\bm{k})p(\bm{x}^n,y^n|\bm{k})\\
&=p(\bm{k})\prod_{i=1}^np(\bm{x}_i,y_i|\bm{x}^{i-1},y^{i-1},\bm{k})\\
&= p(\bm{k})\prod_{i=1}^n \tilde{q}(y_i|\bm{x}_i,\bm{x}^{i-1},y^{i-1},\bm{k})p(\bm{x}_i,\bm{x}^{i-1},y^{i-1},\bm{k})\\
&= \prod_{i=1}^n \tilde{q}(y_i|\bm{x}_i,\bm{x}^{i-1},y^{i-1},\bm{k})\\
&= \prod_{i=1}^n \tilde{q}_{\mathrm{M}_{T,\bm{k}}}(y_i|\bm{x}_i,\bm{x}^{i-1},y^{i-1},\bm{k})
\end{align}
\end{proof}

\newpage
\thispagestyle{myheadings}
\bibliographystyle{junsrt}
\bibliography{master_thesis}

%

%
\end{document}